\documentclass{bmvc2k}
\usepackage{graphicx}
\usepackage{comment}
\usepackage{amsmath,amssymb} 
\usepackage{color}
\usepackage{siunitx}
\usepackage{booktabs}
\usepackage{bm}
\usepackage{mathtools}
\usepackage{xcolor}
\usepackage{pgfplots}
\usepackage{floatrow}
\usepackage{multirow}
\usepackage{wrapfig,lipsum,booktabs}
\usepackage{subcaption}
\usepackage{sidecap}
\usepackage{wrapfig}

\usepackage{xspace}
\usepackage[export]{adjustbox}

\usetikzlibrary{arrows.meta}

\usepackage{bold-extra}

\newcommand*{\eg}{\emph{e.g.}\@\xspace}

\newcommand*{\cf}{\emph{c.f.}\@\xspace}
\newcommand*{\etal}{\emph{et al.}\@\xspace}

\makeatletter
\newcommand*{\etc}{%
    \@ifnextchar{.}%
        {etc.\@\xspace}%
}

\definecolor{ubpubColor}{rgb}{0.43, 0.5, 0.5}
\definecolor{backboneColor}{rgb}{0.423, 0.325, 0.365}
\definecolor{fpnColor}{rgb}{0.255, 0.498, 0.416}

\newcommand{\PAR}[1]{\vskip4pt \noindent {\bf #1~}}

\newcolumntype{P}[1]{>{\centering\arraybackslash}p{#1}}

\newcommand\blfootnote[1]{%
  \begingroup
  \renewcommand\thefootnote{}\footnote{#1}%
  \addtocounter{footnote}{-1}%
  \endgroup
}

\title{Making a Case for 3D Convolutions for Object Segmentation in Videos}

\addauthor{Sabarinath Mahadevan*}{mahadevan@vision.rwth-aachen.de}{1}
\addauthor{Ali Athar*}{athar@vision.rwth-aachen.de}{1}
\addauthor{Aljo\u{s}a O\u{s}ep}{aljosa.osep@tum.de}{2}
\addauthor{Sebastian Hennen}{sebastian.hennen@rwth-aachen.de}{1}
\addauthor{Laura Leal-Taix\'{e}}{leal-taixe@tum.de}{2}
\addauthor{Bastian Leibe}{leibe@vision.rwth-aachen.de}{1}

\addinstitution{
 RWTH Aachen University\\
 Aachen, Germany
}
\addinstitution{
  Technical University of Munich\\
  Munich, Germany
}

\runninghead{Mahadevan, Athar~\etal}{3DC-Seg}

\def\eg{\emph{e.g}\bmvaOneDot}

\def\etal{\emph{et al}\bmvaOneDot}

\begin{document}

\maketitle

\blfootnote{* Equal contribution}

\begin{abstract}
    The task of object segmentation in videos is usually accomplished by processing appearance and motion information separately using standard 2D convolutional networks, followed by a learned fusion of the two sources of information.
    On the other hand, 3D convolutional networks have been successfully applied for video classification tasks, but have not been leveraged as effectively to problems involving dense per-pixel interpretation of videos compared to their 2D convolutional counterparts and lag behind the aforementioned networks in terms of performance.
    In this work, we show that 3D CNNs can be effectively applied to dense video prediction tasks such as salient object segmentation. We propose a simple yet effective encoder-decoder network architecture consisting entirely of 3D convolutions that can be trained end-to-end using a standard cross-entropy loss.
    To this end, we leverage an efficient 3D encoder, and propose a 3D decoder architecture, that comprises novel 3D Global Convolution layers and 3D Refinement modules. Our approach outperforms existing state-of-the-arts by a large margin on the DAVIS'16 Unsupervised, FBMS and ViSal dataset benchmarks in addition to being faster, thus showing that our architecture can efficiently learn expressive spatio-temporal features and produce high quality video segmentation masks.
    We have made our code and trained models publicly available at: \url{https://github.com/sabarim/3DC-Seg}
\end{abstract}

\section{Introduction}
\label{sec:intro}

For a given video clip, the task of segmenting salient objects involves generating binary masks for each frame in that clip, such that all pixels belonging to objects that exhibit dominant or salient motion are labeled as foreground. 
%
%
This is challenging in part because the set of object classes that need to be segmented is not defined a-priori. Therefore, the notion of dominant motion can only be learned by identifying the salient regions based on appearance and motion cues, and capturing the spatial extent and temporal evolution of objects with pixel-level precision over the entire video clip.
This is a core task that is directly related to current problems in computer vision and robotics such video object segmentation~\cite{Perazzi16CVPR,PontTuset17arxiv} and object discovery in videos~\cite{Xiao16CVPR, Osep19ICRA, Wang14ECCV, Kwak15ICCV}.

\begin{figure}[t]
\centering
  \includegraphics[width=0.18\textwidth, frame]{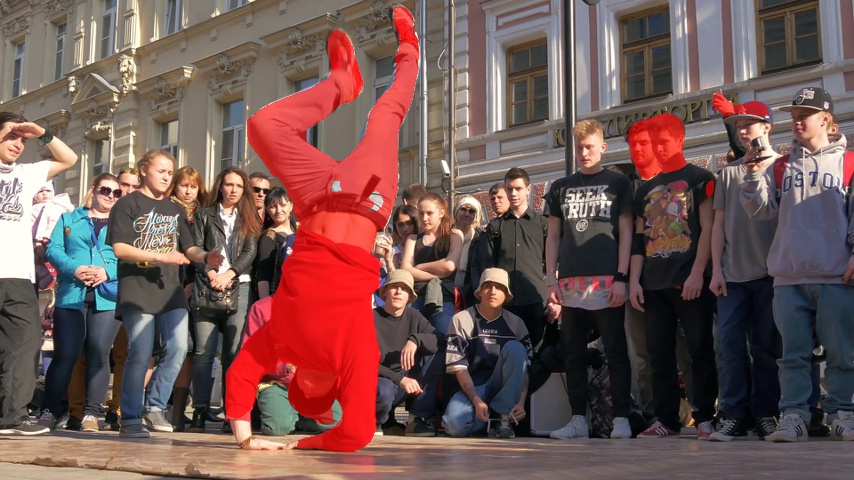}\hspace{1px}%
  \includegraphics[width=0.18\textwidth, frame]{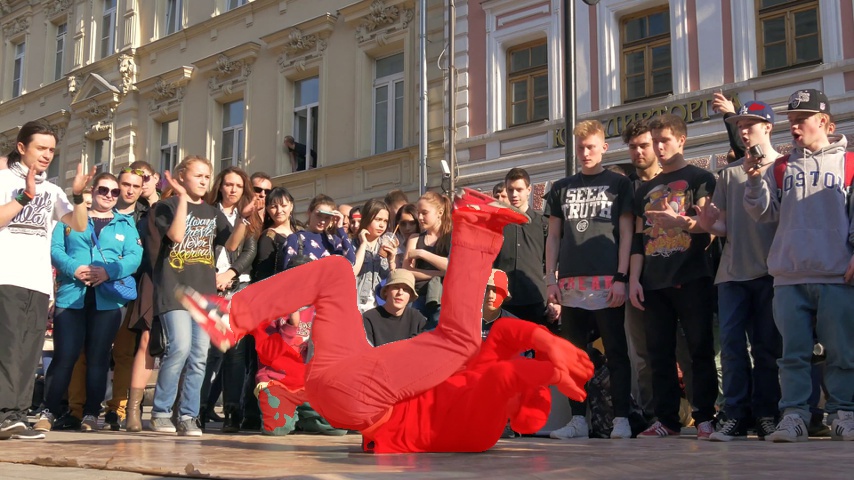}\hspace{1px}%
  \includegraphics[width=0.18\textwidth, frame]{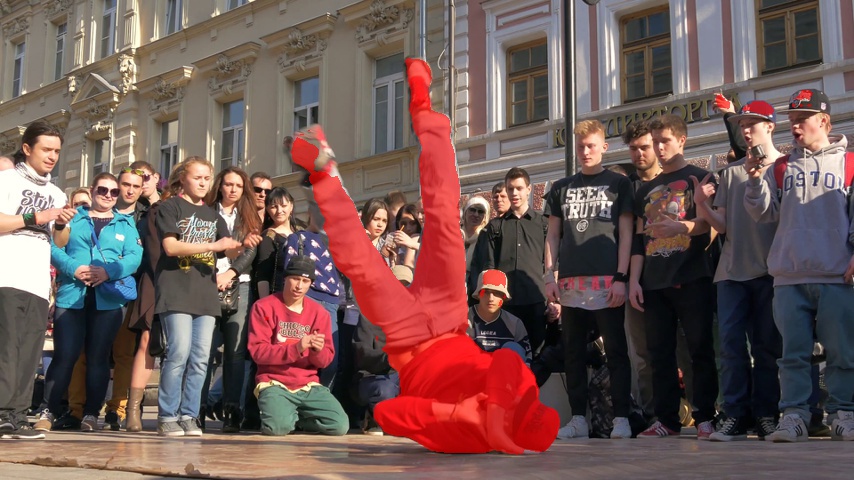}\hspace{1px}%
  \includegraphics[width=0.18\textwidth, frame]{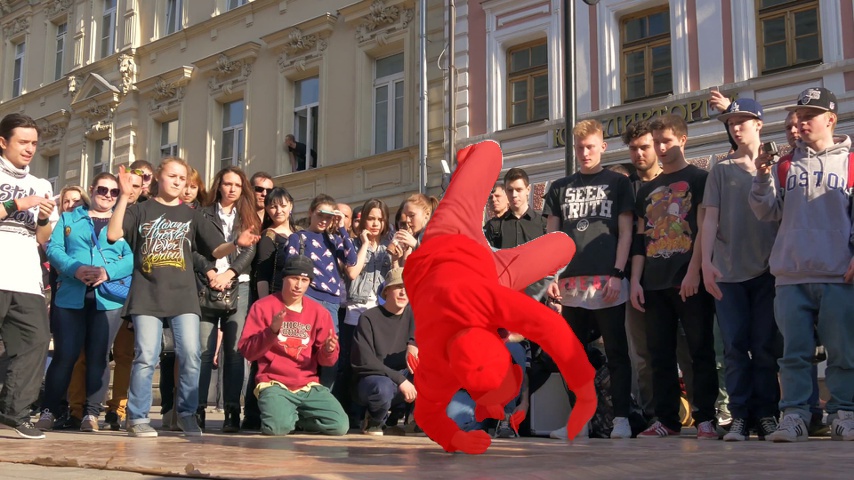}\hspace{1px}%
  \includegraphics[width=0.18\textwidth, frame]{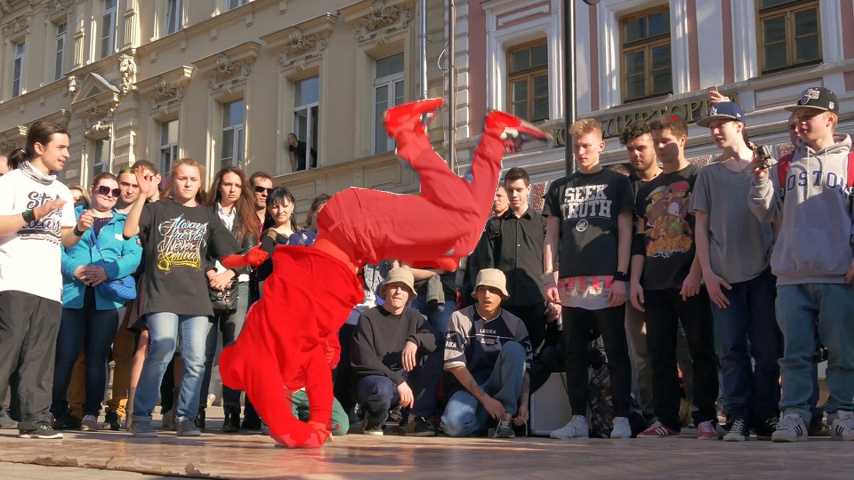}\\
  \includegraphics[width=0.18\textwidth, frame]{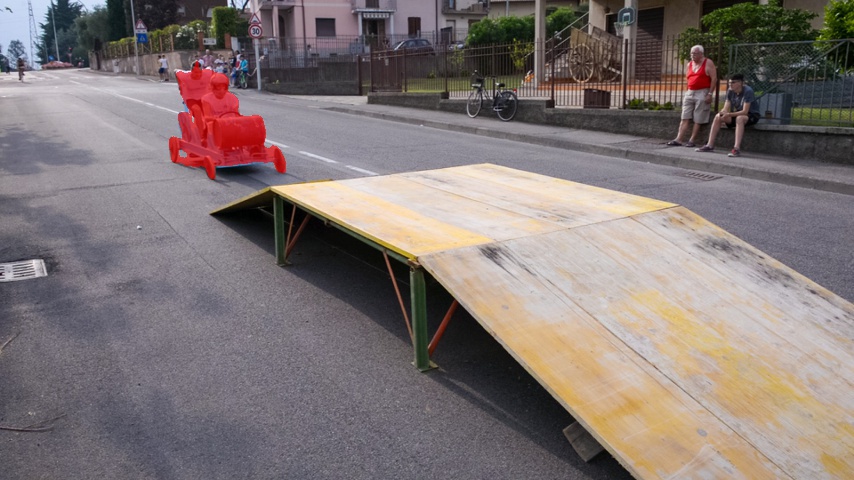}\hspace{1px}%
  \includegraphics[width=0.18\textwidth,frame]{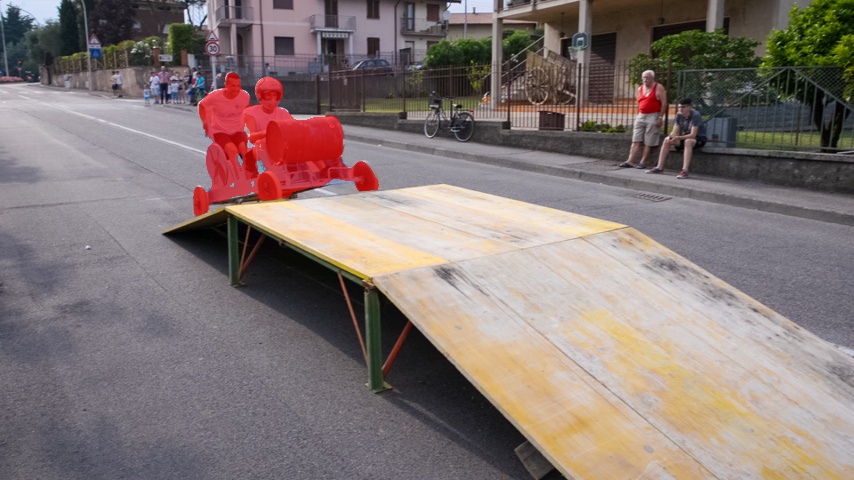}\hspace{1px}%
  \includegraphics[width=0.18\textwidth, trim={0 0 0 0}, clip,frame]{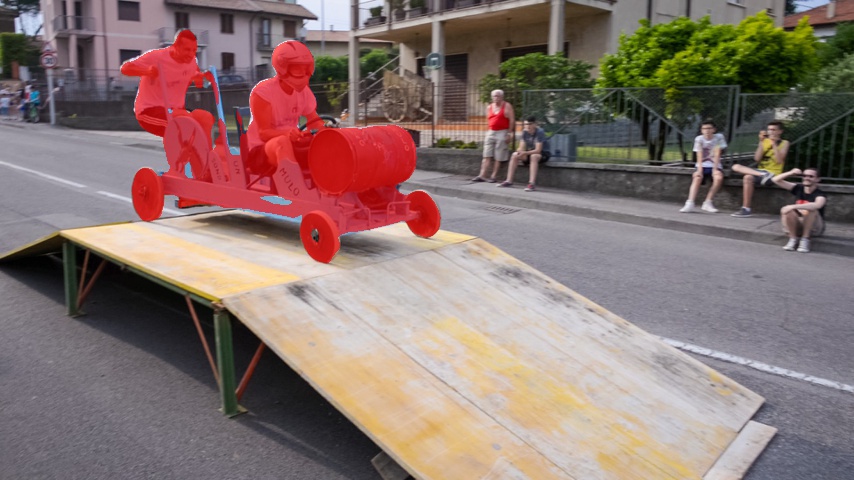}\hspace{1px}%
  \includegraphics[width=0.18\textwidth, trim={0 0 0 0}, clip,frame]{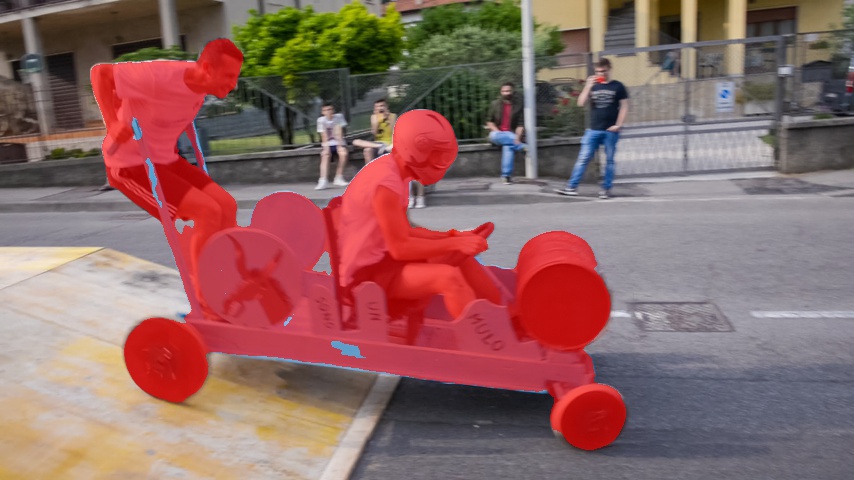}\hspace{1px}%
  \includegraphics[width=0.18\textwidth, trim={0 0 0 0}, clip,frame]{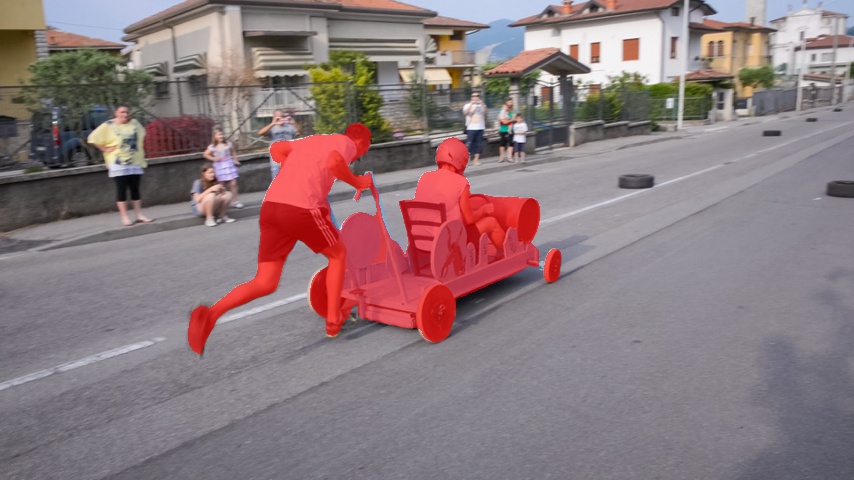}\\
  \includegraphics[width=0.18\textwidth, trim={0 0 0 0}, clip, frame]{figures/teaser/row3/000.jpg}\hspace{1px}%
  \includegraphics[width=0.18\textwidth, trim={0 0 0 0}, clip, frame]{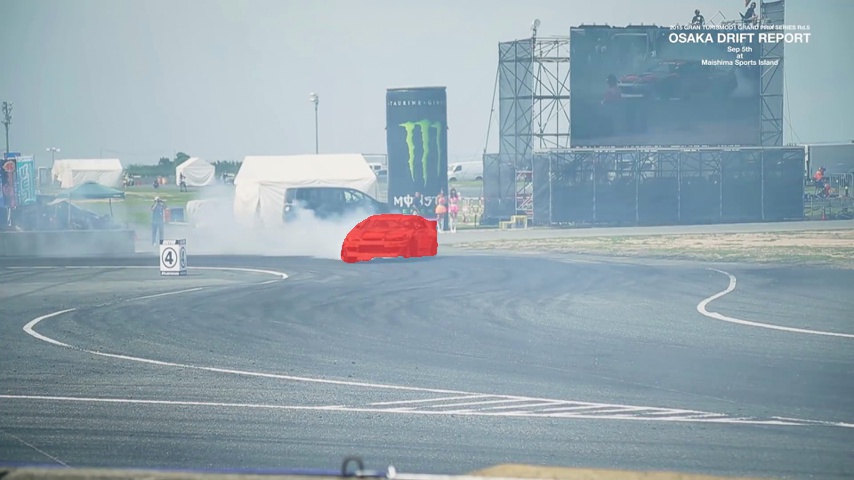}\hspace{1px}%
  \includegraphics[width=0.18\textwidth, trim={0 0 0 0}, clip, frame]{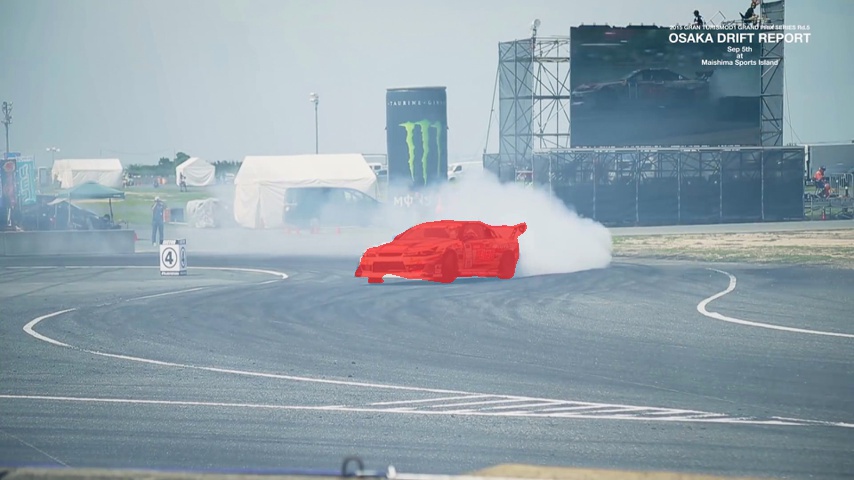}\hspace{1px}%
  \includegraphics[width=0.18\textwidth, trim={0 0 0 0}, clip, frame]{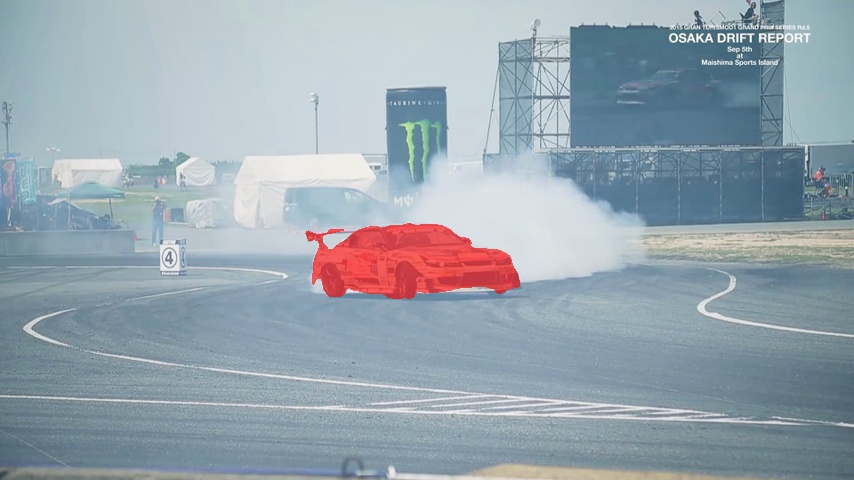}\hspace{1px}%
  \includegraphics[width=0.18\textwidth, trim={0 0 0 0}, clip, frame]{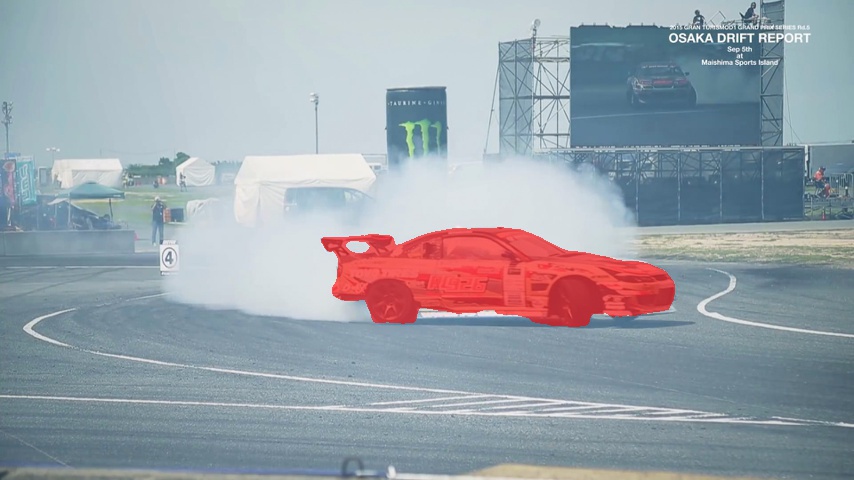}\\
  
\raggedleft
\begin{tikzpicture}[node distance=2cm]
\node (A) at (2.75, 0) {};
\node (B) at (13.0, 0) {};
\draw[-{Stealth}, to path={-- (\tikztotarget)}](A) edge (B);
\node[text width=1cm] at (13.5,0){time};
\end{tikzpicture}
\vspace{-2px}
  \caption{Qualitative results of salient object segmentation masks produced by our network.}
  \label{fig:teaser}
\end{figure}
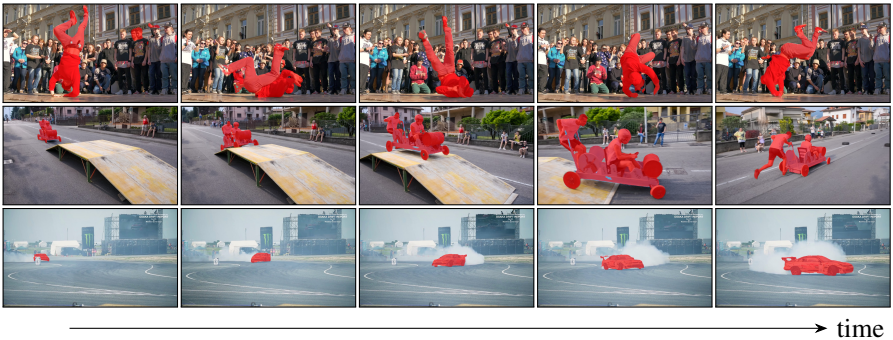

Some existing methods for segmenting salient objects in videos (\eg,~\cite{Jain17CVPR, Tokmakov17ICCV}) follow the approach by \cite{simonyan14nips} and first separately process the appearance and motion information before performing a learned fusion of the two streams of information.
On the other hand, for the task of video action classification, several methods~\cite{Hara18CVPR, Tran18CVPR, Tran19ICCV, Ghadiyaram19CVPR} model videos as 3D volumes and utilize 3D convolutional networks to jointly learn spatial and temporal features and we believe this is a step in the right direction. 
Applying 3D CNNs for pixel-precise segmentation tasks, however, introduces several challenges. Firstly, these networks are generally slower, and contain significantly more trainable parameters compared to their 2D counterparts of the same architecture and depth.
This is especially problematic for segmentation tasks which require higher image resolutions than those used for classification tasks. Secondly, for segmentation tasks it is important (\cf,~\cite{Chen17ARXIV}) to have a network architecture that can capture a large receptive field with respect to each image pixel and effectively leverage multi-scale feature information.

To the best of our knowledge, the work by Hou~\etal~\cite{Hou19BMVC} was the first attempt to apply a fully 3D CNN for video object segmentation.
While promising, it is out-performed by state-of-the-art methods that use 2D CNNs~\cite{Wang19ICCV, Yang19ICCVAnchorDiff}.
This is mainly due to two reasons. Firstly, to keep the computational load manageable, they employ a shallow 3D ResNet-34~\cite{Tran18CVPR} as the encoder network. 
Secondly, they follow the commonly employed design choice of using a smaller stride in the backbone to preserve feature localization, and additionally use atrous convolutions to maintain a large receptive field~\cite{Chen17ARXIV,Chen2018ECCV}. 
As a result, their approach propagates large feature maps through the full depth of the network, which in turn significantly increases the memory footprint and run-time during both inference and training. 

In this paper, we propose a network architecture that mitigates the aforementioned issues and outperforms both \cite{Hou19BMVC} as well as existing state-of-the-art methods based on 2D CNNs~\cite{Jain17CVPR, Tokmakov17ICCV, Yang19ICCVAnchorDiff, Wang19ICCV}. We argue that a better approach for applying 3D CNNs to such tasks involves using a lightweight encoder network with nominal stride. Doing so frees up computational budget which can then be better utilized in enhancing the decoder. In particular, we use a computationally efficient channel-separated network~\cite{Tran19ICCV} pre-trained on large-scale video action classification datasets as the encoder. In the decoder, we use novel 3D variants of Global Convolutions~\cite{Peng17CVPR} and Refinement modules~\cite{Pinheiro16ECCV,Wug18CVPR}. These enable us to capture a large receptive field and learn high-quality segmentation masks from multi-scale encoder features, respectively.
To validate the effectiveness of our network, we apply it to three dataset benchmarks related to salient object segmentation in videos: DAVIS'16 Unsupervised~\cite{Perazzi16CVPR}, Freiburg-Berkeley Motion Segmentation (FBMS)~\cite{Ochs13TPAMI} and ViSal~\cite{Wang15TIP}. We show that our network is both faster than existing state-of-the-arts, and also outperforms them by a large margin. Moreover, we perform several ablation experiments to quantitatively justify our various design choices.

To summarize, in this paper, we (i) demonstrate that 3D CNNs can significantly outperform existing (2D CNN based) methods for tasks involving pixel-precise video segmentation; (ii) propose novel 3D variants of Global Convolutions~\cite{Peng17CVPR} and Refinement modules~\cite{Pinheiro16ECCV,Wug18CVPR} which significantly improve the decoder's performance; and (iii) establish a new state-of-the-art on three datasets.
We believe that our results will motivate others to utilize similar network architectures for other tasks involving pixel-precise video understanding, \eg, discovery of novel object classes~\cite{Xiao16CVPR, Osep19ICRA, Wang14ECCV, Kwak15ICCV}, semi-supervised Video Object Segmentation~\cite{Caelles17CVPR}, Video Instance Segmentation~\cite{Yang19ICCV} and Multi-Object Tracking and Segmentation~\cite{Voigtlaender19CVPR}. 
Fig.~\ref{fig:teaser} shows some qualitative results produced by our network.

\section{Related Work}
\label{sec:related_work}

\PAR{3D CNNs for Video Action Classification}
Early works~\cite{Ji12PAMI,Karpathy14CVPR,Tran15ICCV,Varol17PAMI} which applied 3D CNNs to video human action classification used shallow, often custom network architectures, similar to their 2D CNN counter-parts at the time. 
To overcome the lack of annotated video data, \cite{Carreira17CVPR,Diba18ECCV} proposed novel ways of leveraging 2D image data for training 3D CNNs. 
Later, with the emergence of larger video datasets (\eg~\cite{Kay17ARXIV}), it became possible to effectively train deep 3D CNNs from scratch. \cite{Hara18CVPR} extended the ResNet~\cite{He16CVPR} architecture to 3D by inflating the 3x3 convolutional kernels to 3x3x3. 
Doing so, however, significantly increases the computational overhead. \cite{Xie18ECCV} proposed mixing 2D and 3D convolutions to improve speed and performance whereas \cite{Tran18CVPR} proposed \textit{R(2+1)D} convolutions which factorize 3D convolutions into spatial and temporal convolutions.
Inspired by the success of 2D CNNs with channel-separated convolutions~\cite{Chollet17CVPR}, \cite{Tran19ICCV} proposed a 3D channel-separated ResNet which both performed better and had fewer parameters than existing networks.
\cite{Ghadiyaram19CVPR} improved 3D CNN performance through weakly supervised pre-training on large-scale video data. We show that such pre-training is also beneficial for dense pixel-precise segmentation tasks.

\PAR{Unsupervised Video-Object Segmentation}
The task of \textit{unsupervised} Video Object Segmentation is to estimate a binary segmentation mask for objects in the video clip that exhibit dominant motion.
FusionSeg~\cite{Jain17CVPR} and LVO~\cite{Tokmakov17ICCV} process optical flow and appearance in separate streams before performing a learned fusion of the two.
\cite{Koh17CVPR} generate per-frame object proposals using super-pixels and associate them over time followed by a filtering step to obtain dominant objects.
Different from these works, our 3D CNN approach inherently learns to reason about appearance and motion in a unified manner.
In the same spirit, \cite{Song18ECCV} use Convolutional LSTMs~\cite{xingjian15nips} to leverage the sequential nature of video data and jointly learn spatio-temporal features. However, they use a CRF-based model on top to obtain binary segmentation masks.

In general, methods that employ optical flow, object proposal association, or RNNs struggle with establishing long-range connections. 
To remedy this, AD-Net~\cite{Yang19ICCVAnchorDiff} learns to associate regions of a reference image frame with those in arbitrary query frames. However, such an approach cannot effectively leverage context from several frames.
AGNN~\cite{Wang19ICCV} uses Graph Neural Networks to pass messages between frames in order to model long-range temporal connections.
STEm-Seg~\cite{athar20arxiv} uses an encoder-decoder like architecture with 3D convolutions in the decoder to learn temporal context; however, their encoder network is fully 2D.
\cite{Hou19BMVC} is the most similar to our method as it proposes a fully 3D encoder-decoder network, however, our proposed network architecture differs from theirs and achieves significantly higher performance.

\PAR{Video Salient Object Detection}
Several other works tackle the same problem using various nomenclatures involving video salient object detection. Non deep learning based methods~\cite{Fang14TIP,Wang15CVPR,Wang15TIP,Liu17TCSVT} generally use handcrafted features to create separate intra-frame and inter-frame saliency maps. The task of merging these maps into a coherent sequence of segmentation masks is then formulated as an optimization problem. \cite{Li18CVPR_FGRN} learn a saliency model by using optical flow based motion cues in conjunction with LSTMs, whereas \cite{Wang18TIP} use a CNN to learn single-frame saliency and then apply a dynamic saliency model to handle temporal connections.
Different from all these works, we use 3D CNNs to jointly learn a saliency model over both spatial and temporal domains.

\section{Method}
\label{sec:method}

Our method for segmenting salient object regions in videos is based on an encoder-decoder architecture that leverages 3D convolutions to jointly learn spatio-temporal features. As mentioned in Sec.~\ref{sec:intro}, pixel-precise segmentation tasks benefit from higher image resolutions and networks with large receptive fields, which is computationally challenging when working with 3D CNNs.
%
%
%
We mitigate these challenges by employing an efficient channel-separated encoder network~\cite{Tran19ICCV}, and a decoder comprising (i) novel 3D Global Convolutions (GC3D) which can capture a large receptive field, and (ii) novel 3D Refinement modules which effectively refine multi-scale encoder features into high quality segmentation masks.

\subsection{Encoder}
\label{sec:encoder}

\begin{wraptable}{r}{0.5\textwidth}
\centering
\scriptsize
\sisetup{detect-weight=true}
\setlength{\tabcolsep}{5px}%
\vspace{-10pt}
\begin{tabular}{|l|c|c|c|}
\toprule
 \multicolumn{4}{c}{Backbone Comparison} \\
 \midrule
 Architecture & Type & \# Params            & Runtime \\
              &      & $\left(\times 10^6 \right)$ & (sec)     \\
\cmidrule(lr){1-4}
ResNet-101~\cite{He16CVPR}                     & 2D & 42.5 & 0.173 \\
DeepLabV3 ResNet-101~\cite{Chen17ARXIV}        & 2D & 42.6 & 0.793 \\
ResNet-34 R2+1D $^*$~\cite{Tran18CVPR} & 3D & 63.5 & 0.891 \\
ResNet-152 (ir-CSN)~\cite{Tran19ICCV}    & 3D & 28.7 & 0.213 \\
\bottomrule
\end{tabular}
\caption{Comparison of various backbones. Runtime is for generating feature maps for an 8-frame clip with $854\times 480$ resolution on an Nvidia GTX-1080Ti. $^*$: lower stride~\cite{Hou19BMVC}} 
\vspace{-2mm}
\label{tab:backbone_comparison}
\end{wraptable}

The encoder of our network is a computationally efficient 3D ResNet with channel-separated convolutions which has been successfully used for video action classification~\cite{Tran19ICCV}. In particular, we use the reduced interaction (ir-CSN) variant of their model in which every 3x3x3 convolution in the \textit{bottleneck} block of the ResNet is replaced with a 3x3x3 depth-wise separable convolution, while the pre-existing 1x1x1 convolutions in the \textit{bottleneck} block capture channel interactions. The reduced memory footprint of this architecture enables a 152 layer variant of this network to be feasibly applied to pixel-precise segmentation tasks in conjunction with our proposed decoder architecture (Sec.~\ref{sec:decoder}). 
%
%
To justify this design decision, we provide a quantitative analysis of the computational overhead of various backbones used in recent works in Tab.~\ref{tab:backbone_comparison}. 
Despite being a \textit{deeper backbone}, the ResNet-152 based ir-CSN has significantly fewer parameters compared to other shallower networks. In terms of runtime, only the vanilla 2D ResNet-101 is slightly faster, however, such 2D networks are inherently unable to learn temporal context. 

State-of-the-art methods~\cite{Yang19ICCVAnchorDiff,Wang19ICCV} either employ a 2D network such as DeepLabV3's~\cite{Chen17ARXIV} ResNet-101 backbone, or a shallow 3D network with atrous convolutions and reduced stride \cite{Hou19BMVC}. Though this strategy improves performance in segmentation tasks, a major drawback is that it also significantly increases the memory footprint and run-time. We argue that a better approach is to use a computationally efficient channel-separated backbone with nominal stride. Not only does this enable us to have a deeper backbone which can generally learn better features for the end-task, but more importantly, it frees up valuable computational budget that can be used to enhance the decoder's efficacy.


\subsection{Decoder}
\label{sec:decoder}

For an input video clip, the encoder produces feature maps at 4 different scales. The decoder architecture comprises a series of 3D convolutions and up-sampling layers which refine these feature maps into the final segmentation mask. 
To capture a large receptive field in the encoder features, Chen \etal~\cite{Chen2018ECCV} proposed using an encoder with reduced stride (8x or 16x) in combination with an Atrous Spatial Pyramid Pooling (ASPP) module which applies multiple parallel atrous convolutions with different dilation rates to a feature map. \cite{Hou19BMVC} also proposed a 3D variant of ASPP and used it in their network. 

\begin{wrapfigure}[18]{r}{0.5\textwidth}
\centering
    \includegraphics[width=0.9\textwidth]{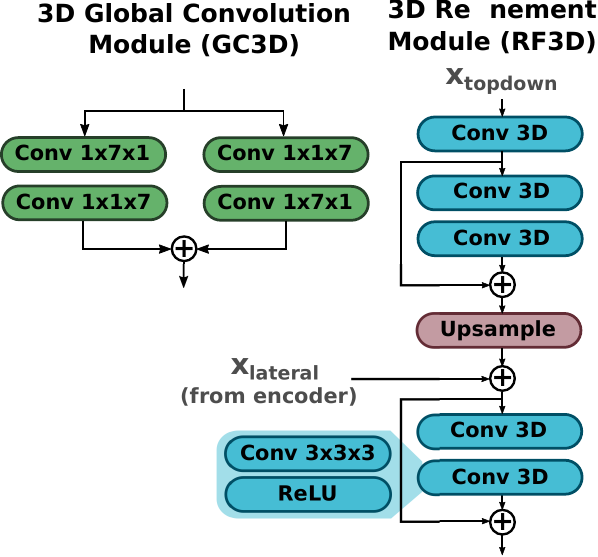}
    \caption{3D Global Convolution (GC3D) and 3D Refinement modules (RF3D) illustrated.}
    \label{fig:modules}
\vspace{-3mm}
\end{wrapfigure}
%
%
By contrast, we learn encoder features at the nominal 32x stride commonly used for classification tasks.
To capture wide spatial context, we propose a 3D variant of \textit{Global Convolutional Networks} which were introduced for semantic segmentation in images~\cite{Peng17CVPR}.
The idea here is that a large $k\times k$ convolution can be replaced with a series of row and column convolutions with kernel sizes $1\times k$ and $k\times 1$, respectively. This yields the same effective receptive field while having fewer parameters. 
Our 3D Global Convolution module (GC3D) comprises 3D convolutions with unity kernel size along the temporal dimension. This is because the temporal dimension of the input video clip is usually much smaller than the spatial dimensions.

\begin{figure}
    \centering
    \includegraphics[width=\textwidth]{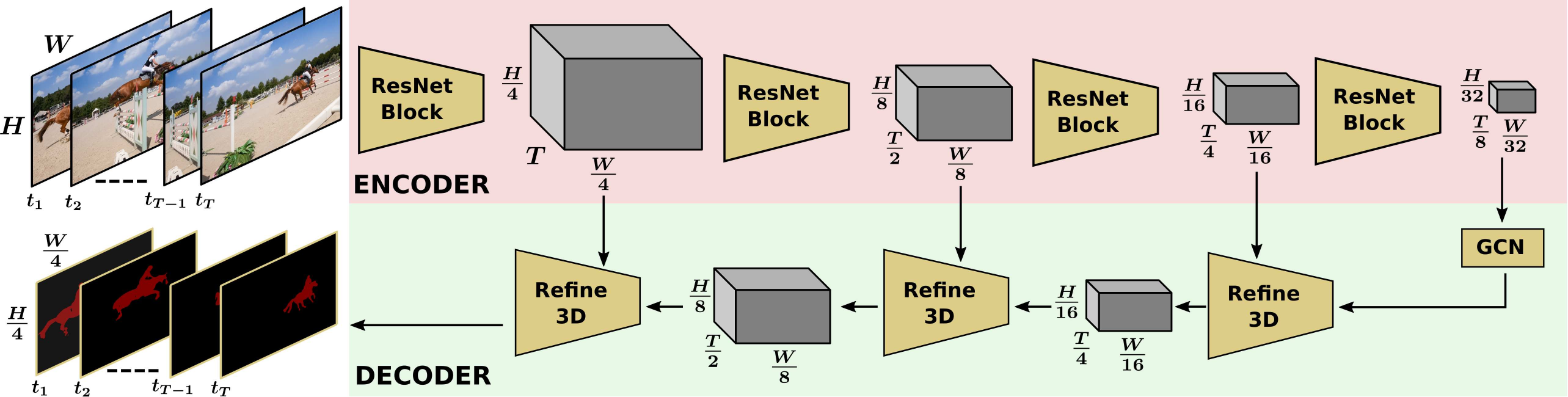}
    \caption{Illustration of our encoder-decoder network architecture}
    \label{fig:network_architecture}
\end{figure}

To combine and upsample the multi-scale feature maps, we additionally propose a 3D variant of the \textit{Refinement module} introduced in \cite{Pinheiro16ECCV} for object proposal generation in images. 
The basic idea here is to apply two 3x3x3 convolutions to a given feature map with a skip connection, followed by trilinear upsampling and addition with the corresponding encoder feature map at that scale. This is followed by two further convolutions with a skip connection. Both GC3D and 3D Refinement modules are illustrated in Fig.~\ref{fig:modules}, and the overall network architecture is illustrated in Fig.~\ref{fig:network_architecture}.

\subsection{Video Clip Sampling}
\label{sec:clip_sampling}

For optimal network performance, the input clip's temporal length should be consistent between training and inference. Therefore, to apply the network to videos of arbitrary length, we divide the input video into clips of length $T_c$ with an overlap of $T_o$ between successive clips. For overlapping frames, the mask probabilities are averaged to produce the final segmentation masks. Generally, our method is therefore \textit{near-online}, because given a new frame, the segmentation mask for it becomes available after at most $T_c-T_o-1$ time-steps (except for the very first $T_c$ frames in the video stream). Note that an online variant can be realized if $T_o \gets T_c-1$.



\section{Experiments}
\label{sec:experiments}

\subsection{Datasets and Evaluation}

\PAR{DAVIS'16.} Densely Annotated Video Instance Segmentation (DAVIS) is a popular set of benchmarks for video object segmentation related tasks. We evaluate on the DAVIS'16 \textit{unsupervised} benchmark~\cite{Perazzi16CVPR} which contains 30 videos for training and 20 for validation/testing. 
The task is to produce a segmentation mask that captures the dominant objects in the video. 
Note that the \textit{unsupervised} task differs from the more popular \textit{semi-supervised} task in which ground-truth annotations of the first frame are known during inference. The evaluation metrics used are (i) $\mathcal{J}$-mean, which is the intersection-over-union of the predicted and ground truth masks, and (ii) $\mathcal{F}$-mean, which measures the accuracy of the predicted mask boundaries. These measures can be averaged to give an overall $\mathcal{J}\&\mathcal{F}$ score.

\PAR{FBMS.} The Freiburg-Berkeley Motion Segmentation dataset~\cite{Ochs13TPAMI} contains 59 videos which include 12 videos from the Hopkins-155 dataset~\cite{Tron07CVPR}. 
The ground truth annotation for every 20\textsuperscript{th} frame is provided resulting in a total of 720 annotated frames in the entire dataset.

\PAR{ViSal.} The Video Saliency dataset~\cite{Wang15TIP} is a collection of 17 videos with a diverse set of objects and backgrounds, varying in length from 30 to 100 frames.
For both FBMS and ViSal, the evaluation measures are (i) F-measure, which is the harmonic mean of the per-pixel precision and recall scores, and (ii) the Mean Absolute Error (MAE) $\in [0,1]$ between the predicted and ground truth segmentation masks.

\subsection{Training}
\label{sec:training}

As mentioned in Sec.~\ref{sec:encoder}, our encoder is an \textit{ir-CSN} variant of a 3D ResNet-152. We initialize its weights from a model pre-trained on IG-65M~\cite{Ghadiyaram19CVPR} and Kinetics~\cite{Kay17ARXIV}. The decoder weights are initialized randomly. 
The network's inputs are video clips of length $T_c=8$ (the same clip length is used for inference). We sample training clips on the fly from a video sequence of length $L$ by first choosing a random frame $t \in \{1,..., L\}$, and then sampling the remaining $T_c-1$ frames randomly from $\{t+1,\ldots, \text{min}(t+S, L)\}$.
Here, $S=32$ is a hyper-parameter which limits the maximum temporal span of a training clip. If $t+T_c < L$, the video is padded with its last frame until $t+T_c = L$. 

The network is trained end-to-end using the Adam~\cite{Kingma15ICLR} optimizer with an initial learning rate of $10^{-5}$ which is decayed exponentially after every epoch. The network is first trained on synthetically generated video clips from the COCO Instance Segmentation dataset~\cite{Lin14ECCV}, followed by a second stage with video data from the YouTube-VOS~\cite{Xu18ECCV} and DAVIS'16~\cite{Perazzi16CVPR} datasets. 

\PAR{Pre-training on Images:} Similar to~\cite{athar20arxiv,Wug18CVPR,Wang18TIP}, our network generalizes better if we train on static image datasets in addition to video data. For this, we synthesize video clips from static images by augmenting them using a combination of random affine and piecewise-affine transformations. These transformations are applied sequentially to mimic video motion.
To obtain the ground truth masks, we combine all object instances into a single foreground mask before applying the same set of transformations.
    
\subsection{Ablations}

\begin{table}[t]
    \centering
    \footnotesize
        \begin{subtable}[t]{0.65\textwidth}
        {
            \begin{tabular}[t]{c| c | c | c }
\toprule
             {\footnotesize Backbone} & Pre-training & Fine-tuning & $\mathcal{J}$\&$\mathcal{F} (\%)$  \\
            \cmidrule(lr){1-4}
                            ir-CSN-152 & Sports1M + Kinetics & COCO& 61.8 \\
                            ir-CSN-152 & Sports1M + Kinetics & DAVIS & \textcolor{blue}{82.2} \\
                            ir-CSN-152 & Sports1M + Kinetics & YT-VOS&
                            69.5 \\
                            ir-CSN-152 & Sports1M + Kinetics &COCO, YT-VOS& 72.6 \\
                            ir-CSN-152 & Sports1M + Kinetics &YT-VOS, DAVIS& 79.4 \\
                            ir-CSN-152 & Sports1M + Kinetics &COCO, DAVIS& \textcolor{blue}{82.8} \\
            \cmidrule(lr){1-4}
                            ir-CSN-152 &Sports1M + Kinetics &COCO, YT-VOS, DAVIS& \textcolor{blue}{83.6}\\
                            ir-CSN-152 &IG-65M + Kinetics &COCO, YT-VOS, DAVIS& \textcolor{blue}{84.1}\\
            \cmidrule(lr){1-4}
                            R(2+1)D-34 &IG-65M + Kinetics& COCO, YT-VOS, DAVIS & 79.6\\
            \bottomrule
            \end{tabular}
            \caption{}
            \label{tab:ablation_data}
        }
        \end{subtable}
        \hfill
        \begin{subtable}[t]{0.25\linewidth}%
        {
            \begin{tabular}[t]{lr}%
            {\scriptsize\textbf{(b)}}&%
            \hspace{-3mm}
            \begin{subtable}[t]{\linewidth}
            {
            \begin{tabular}[t]{l | c} 
                \toprule
                Module & $\mathcal{J}\&\mathcal{F}$ \\
                \cmidrule(lr){1-2}
                    C3D  &  80.3\\
                    NL3D  & \textcolor{blue}{81.7}\\
                    ASPP & 81.0\\
                    GC3D   & \textcolor{blue}{84.1}\\
                \bottomrule
                \end{tabular}
                \phantomsubcaption{}
                \label{tab:ablation_modules}
            }
            \end{subtable}\\
            {\scriptsize\textbf{(c)}}&%
            \hspace{-3mm}
            \begin{subtable}[t]{\linewidth}
            {
            \begin{tabular}[t]{l | c} 
                \toprule
                Temporal \\Length & $\mathcal{J}\&\mathcal{F}$ \\
                \cmidrule(lr){1-2}
                    4  &  \textcolor{blue}{83.4}\\
                    8 & \textcolor{blue}{84.1}\\
                    16   & \textcolor{blue}{83.7}\\
                    24   & 81.5\\
                \bottomrule
                \end{tabular}
                \phantomsubcaption{}{}
                \label{tab:ablation_tw}
            }
        \end{subtable}
            \end{tabular}
        }
        \end{subtable}
        \hfill
        \caption{Ablation studies on DAVIS'16 \texttt{val}:  (\subref{tab:ablation_data}): Comparison of different backbones and the impact of training data; (\subref{tab:ablation_modules}): Effect of bridging modules; (\subref{tab:ablation_tw}): Performance study on different temporal window size. C3D: 3D convolution, NL3D: Non-local 3D, GC3D: 3D Global convolution. Scores higher than the existing state-of-the-art are highlighted in \textcolor{blue}{blue}.}
        \label{tab:ablation}
\end{table}

We perform several ablation experiments on the DAVIS'16 unsupervised validation set to justify our design choices.

\PAR{Backbone and Training Data:}
Tab.~\ref{tab:ablation_data} shows the $\mathcal{J\&F}$ scores for different encoder networks and the corresponding datasets used for (pre-)training. It can be seen that our model performs consistently well regardless of the encoder network depth and the amount of (pre-)training data.
Using a shallow ResNet-34 \textit{R(2+1)D} encoder network~\cite{Tran18CVPR} (last row), our network achieves 79.6\% $\mathcal{J}\&\mathcal{F}$. This is only 1.5\% behind the current state-of-the-art method AD-Net~\cite{Yang19ICCVAnchorDiff} (see Tab.~\ref{tab:davis16-bench}) which uses a low stride ResNet-101 backbone from DeepLabV3~\cite{Chen17ARXIV} in addition to heuristic post-processing.

In the interest of comparing against existing 2D CNN approaches which typically use ImageNet~\cite{Deng09CVPR} pre-training, we conducted several ablations using Sports1M~\cite{Karpathy14CVPR} + Kinetics~\cite{Kay17ARXIV} pre-trained weights. 
The number of data samples in Sports1M + Kinetics (1.8M video clips) is comparable to ImageNet (1.2M images). However, we stress that a direct comparison is difficult since (i) the video datasets have more images, but unlike ImageNet, the image frames within a video are highly correlated, and (ii) ImageNet contains 1000 highly diverse object classes, but Kinetics and Sports-1M are restricted to human action classes. 
With this setting, fine-tuning on DAVIS alone yields 82.2\% $\mathcal{J\&F}$ (row 2) which outperforms the existing state-of-the-art by 1.1\% (see Tab.~\ref{tab:davis16-bench}). However, since DAVIS is a small dataset with only 30 training sequences, we obtained further improvements by training on multiple datasets. With additional static image training using COCO~\cite{Lin14ECCV}, as in the (COCO, DAVIS) setting, the score improves to 82.8\% (row 6). By further adding YouTube-VOS~\cite{Xu18ECCV} to the training set (COCO, DAVIS, YT-VOS), the score improves to 83.6\% (row 7). Finally, using pre-trained weights from the much larger IG-65M dataset~\cite{Ghadiyaram19CVPR} and fine-tuning on all three datasets yields the best score of 84.1\% (row 8).

We conclude that even though using more (pre-)training data improves performance, our model achieves state-of-the-art scores even with training data settings that are comparable to existing methods. Secondly, the efficacy of static image training is evident from the fact that the (COCO, DAVIS) setting yields 82.8\% $\mathcal{J\&F}$ which is 3.4\% higher than the 79.4\% obtained with video data only (YT-VOS, DAVIS). For the sake of completeness, we also report results with models fine-tuned only on COCO (row 1) and YT-VOS (row 3).


\PAR{Decoder:} To justify our decoder architecture, we ablate its two major components: the 3D Global Convolution (GC3D) and 3D Refinement modules.  
In our network, the GC3D module is applied to the final (smallest) output feature map of the encoder to capture a large receptive field. 
In Tab.~\ref{tab:ablation_modules}, we compare the network's performance when the GC3D module is replaced by (i) a Non-Local 3D (NL3D) block~\cite{Wang17CVPR}, (ii) an Atrous Spatial Pyramid Pooling (ASPP) module, and (iii) a vanilla 3x3x3 convolution baseline (C3D). It can be seen that ASPP (81.0\%) outperforms C3D (80.3\%) by 0.7\% $\mathcal{J\&F}$, and NL3D further improves this by another 0.7\% (81.7\%), but GC3D outperforms all of these modules (84.1\%) yielding a 3.8\% improvement over the baseline C3D. This highlights the effectiveness of using the GC3D module in our network.

\begin{wraptable}{r}{0.25\textwidth}
\setlength{\tabcolsep}{5px}%
\footnotesize
\vspace{-19pt}%
\begin{tabular}[t]{l | c} 
            \toprule
            Module & $\mathcal{J}\&\mathcal{F}$ \\
            \cmidrule(lr){1-2}
                Upsampling &80.2 \\
                RF3D  &  84.1\\
            \bottomrule
            \end{tabular}
\caption{Analysis of different decoder modules on DAVIS'16.}
\vspace{-8px}
\label{tab:ablation_refinement}
\end{wraptable}

The second major component of our decoder is the 3D Refinement module which helps the network in recovering the spatial and temporal resolution from the feature maps generated by the encoder. Tab.~\ref{tab:ablation_refinement} compares our 3D refinement module (RF3D) against a baseline \textit{Upsampling} block which contains two 3x3x3 convolutions followed by a concatenation with encoder features and subsequent trilinear upsampling. As it can be seen, the 3D Refinement module (84.1\%) improves performance on DAVIS'16 by 3.9\% $\mathcal{J\&F}$ compared to the \textit{Upsampling} baseline (80.2\%), thereby showing its effectiveness in recovering the spatio-temporal resolution. 

\PAR{Input Clip Length:} Finally, we ablate the effect of varying the input clip length ($T_c$) and report the results in Tab.~\ref{tab:ablation_tw}. \cite{Tran18CVPR} noted that 3D CNNs can be initially trained with a lower $T_c$ followed by fine-tuning on the target $T_c$ without sacrificing performance. Following this, we first train with $T_c=8$ on COCO and YouTube-VOS, followed by fine-tuning with the reported $T_c$ on DAVIS. 
As can be seen, our method is robust to large variations of $T_c$ between 4 and 16. For $T_c>16$ however, the performance decreases. This highlights our architecture's limitation in coping with very large temporal dimensions, which we leave for future work.

\subsection{Benchmark results}

\begin{table}[t!]
\footnotesize
\setlength{\tabcolsep}{4pt} %
\centering
\sisetup{detect-weight=true}
\begin{tabular}{l|ccc |ccc|c}
\toprule
 \multicolumn{8}{c}{DAVIS 2016 Unsupervised} \\
 \midrule
 Method & OF & CRF & MS & $\mathcal{J}$\&$\mathcal{F}$ & $\mathcal{J}$-mean& $\mathcal{F}$-mean & Time (s/frame)\\
\cmidrule(lr){1-8}
OnAVOS~\cite{Voigtlaender17BMVC} & & & &- & 72.7 & - & - \\
ARP~\cite{Koh17CVPR} & \checkmark &&  & 73.4 & 76.2& 70.6 & - \\
LVO~\cite{Tokmakov17ICCV} & \checkmark & \checkmark && 74.0 & 75.9 & 72.1 & - \\
PDB~\cite{Song18ECCV} & &\checkmark&& 75.9 & 77.2 & 74.5 & - \\
MotAdapt~\cite{SiamICRA2019}  & &&& 77.3 & 77.2 & 77.4 & - \\
3D-CNN~\cite{Hou19BMVC}  &&&& 77.8 & 78.3 & 77.2 & 0.38 \\
AD-Net~\cite{Yang19ICCVAnchorDiff} & & & \checkmark & 78.8 & 79.4 & 78.2 & 0.38 \\
AGNN~\cite{Wang19ICCV}  & &\checkmark&\checkmark& 79.9 & 80.7 & 79.1 & 2.96\\
STEm-Seg~\cite{athar20arxiv}  & &&\checkmark& 80.6 & 80.6 &80.6 & 1.42\\
AD-Net + Inst-Pruning~\cite{Yang19ICCVAnchorDiff}$^*$  & &&\checkmark& 81.1 & 81.7& 80.5 & 2.94\\
\cmidrule(lr){1-8}
Ours  & &&& 84.1 & 83.9 &84.3 & \textbf{0.22}\\
Ours - Dense  & &&& \textbf{84.5} & \textbf{84.3} & \textbf{84.7} & 0.84\\
\bottomrule
\end{tabular}
\caption{\label{tab:evaltable2} DAVIS'16 validation set results for the unsupervised track. OF: Optical Flow, MS: Multi-Scale inference, CRF: CRF post-processing. Runtime was computed on an Nvidia GTX-1080Ti. $^*$ Uses heuristic post-processing. Best performance scores are highlighted in bold.}
\vspace{-2mm}
\label{tab:davis16-bench}
\end{table}

\PAR{DAVIS 2016 Unsupervised:} We apply our network to the DAVIS'16 unsupervised benchmark~\cite{Perazzi16CVPR} and report the results in Tab.~\ref{tab:davis16-bench}.
Our 3D CNN achieves 84.1\% $\mathcal{J}\&\mathcal{F}$, which is a substantial 3\% improvement over the existing state-of-the-art of 81.1\%. It also performs better in terms of the individual $\mathcal{J}$-mean and $\mathcal{F}$-mean measures.
This shows that our no-bells-and-whistles encoder-decoder network is able to produce high quality segmentation masks by jointly learning the salient objects' appearance and motion models. 
By contrast, several competing methods perform inference at multiple input scales and/or apply post-processing techniques such as CRFs to improve performance which imposes additional computational overhead and renders the method non-end-to-end trainable. 
The best performing existing method, AD-Net~\cite{Yang19ICCVAnchorDiff}, applies an \textit{instance pruning} post-processing step that additionally requires inference with a separate image instance segmentation network which is trained on COCO~\cite{Lin14ECCV}. Under this setting, AD-Net uses ImageNet and DAVIS for (pre-)training its primary network and COCO~\cite{Lin14ECCV} for post-processing. Without this post-processing step, AD-Net achieves 79.4\% $\mathcal{J}\&\mathcal{F}$ which is 4.7\% lower than our score of 84.1\%. STEm-Seg~\cite{athar20arxiv}, the second-best performing existing method, uses a backbone network initialized from Mask-RCNN~\cite{He17ICCV} weights, and then further fine-tunes jointly on COCO~\cite{Lin14ECCV}, YouTube-VIS~\cite{Yang19ICCV} and DAVIS~\cite{Perazzi16CVPR}. It also employs multi-scale inference whereas we do not.
The existing 3D CNN approach~\cite{Hou19BMVC} fine-tunes only on DAVIS and achieves 77.8\% $\mathcal{J}\&\mathcal{F}$ which is 3.4\% lower than our comparable ablation score of 82.2\% (see Tab.~\ref{tab:ablation_data}, row 2).
Due to our decision of using a nominal stride and an efficient encoder network, our method is also the fastest among recent works. It runs at 0.22 s/frame (4.5fps) on an Nvidia GTX-1080Ti which is 42\% faster than the two tied second fastest methods (\textit{AD-Net} and \textit{3D-CNN} with 0.38 s/frame or 2.6fps).

We use a frame overlap of $T_o = 3$, however, as mentioned in Sec.~\ref{sec:clip_sampling}, an online version can be realized if $T_o \gets T_c-1$. The results for this setting are given as \textit{Ours - Dense} and the performance is slightly better, but this setting is slower.

\PAR{Video Object Saliency:} This task involves segmenting pixels in a video which belong to \textit{salient} objects, and is similar to unsupervised video object segmentation. Our method can therefore be directly evaluated on it. 

\begin{wraptable}{r}{0.6\textwidth}
\vspace{-5mm}
\footnotesize
\setlength{\tabcolsep}{4pt} %
\centering
\sisetup{detect-weight=true}
\begin{tabular}{l|cc |cc|cc}
\toprule
 &\multicolumn{2}{c|}{DAVIS '16} & \multicolumn{2}{c|}{FBMS} & \multicolumn{2}{c}{ViSal} \\
 \cmidrule(lr){2-7}
 Method & F$\uparrow$ & MAE$\downarrow$ & F$\uparrow$ & MAE$\downarrow$ & F$\uparrow$ & MAE$\downarrow$ \\
\cmidrule(lr){1-7}
FGRNE~\cite{Li18CVPR} & 78.6 & 0.043 & 77.9 & 0.083& 85.0 & 0.040\\ 
FCNS~\cite{Wang17TIP} & 72.9 & 0.053& 73.5& 0.100& 87.7& 0.041 \\
SGSP~\cite{Liu16TCSVT} & 67.7& 0.128& 57.1&  0.171& 64.8&  0.172 \\
GAFL~\cite{Wang15TIP} & 57.8 & 0.091& 55.1&  0.150& 72.6&  0.099 \\
SAGE~\cite{Wang15CVPR} & 47.9& 0.105& 58.1&  0.142& 73.4&  0.096 \\
STUW~\cite{Fang14TIP} & 69.2& 0.098& 52.8&  0.143& 67.1& 0.132\\
SP~\cite{Liu14TCSVT}   & 60.1&0.130&53.8&0.161&73.1&0.126\\
AD-Net~\cite{Yang19ICCVAnchorDiff} & 80.8 &0.044  &81.2 & 0.064  &90.4 &0.030 \\
\cmidrule(lr){1-7}
Ours&  \textbf{91.8} &\textbf{0.015}& \textbf{84.5} & \textbf{0.048} & \textbf{92.2} & \textbf{0.019}\\
\bottomrule
\end{tabular}
\caption{\label{tab:evaltable2} F-measure and MAE for DAVIS, FBMS and ViSal datasets. $\uparrow$: Higher is better, $\downarrow$: Lower is better.}
\label{tab:saliency}
\vspace{-2mm}
\end{wraptable}

In addition to evaluating our DAVIS'16~\cite{PontTuset17arxiv} results using the F-measure and MAE, we also evaluate on the FBMS~\cite{Ochs13TPAMI} and ViSal~\cite{Wang15TIP} datasets without any additional dataset-specific training. The scores for all three datasets are reported in Tab.~\ref{tab:saliency}. Our method outperforms the state-of-the-art on all these datasets for both evaluation measures, thus signifying its generalization capability. Note that the performance improvement on DAVIS'16 is particularly high compared to the second-best method.

\section{Conclusion}

In this paper, we proposed a simple and fast network architecture consisting entirely of 3D convolutions that is capable of effectively learning spatio-temporal features without additional bells and whistles. To this end, we employed a deep yet computationally efficient 3D ResNet pretrained for video action classification as an encoder, and a novel decoder architecture inspired by existing 2D convolutional networks. 
Our experiments show that in addition to being faster than existing state-of-the-art methods, our network also out-performs them on three different datasets by a large margin. 
We believe that our findings will encourage other researchers to employ 3D convolutions for a variety of tasks involving pixel-precise video scene understanding, and that our proposed network architecture can serve as a useful starting point for their endeavours.

\PAR{Acknowledgements.}
This project was funded, in parts, by ERC Consolidator Grant DeeVise (ERC-2017-COG-773161), EU project CROWDBOT
(H2020-ICT-2017-779942) and the Humboldt Foundation through the Sofja Kovalevskaja
Award. Computing resources for several experiments were granted by RWTH Aachen University under project 'rwth0519'. We thank Paul Voigtlaender and Istv{\'a}n S{\'a}r{\'a}ndi for helpful discussions.

\clearpage
\bibliography{abbrev_short,mybib}

\clearpage
\appendix
\textcolor{bmv@sectioncolor}{\vspace{-1.2cm} \part*{Supplementary Material}}
\section{Qualitative Results}

\vspace{-8pt}
\begin{figure}[h]
\centering
\includegraphics[width=0.18\textwidth, frame]{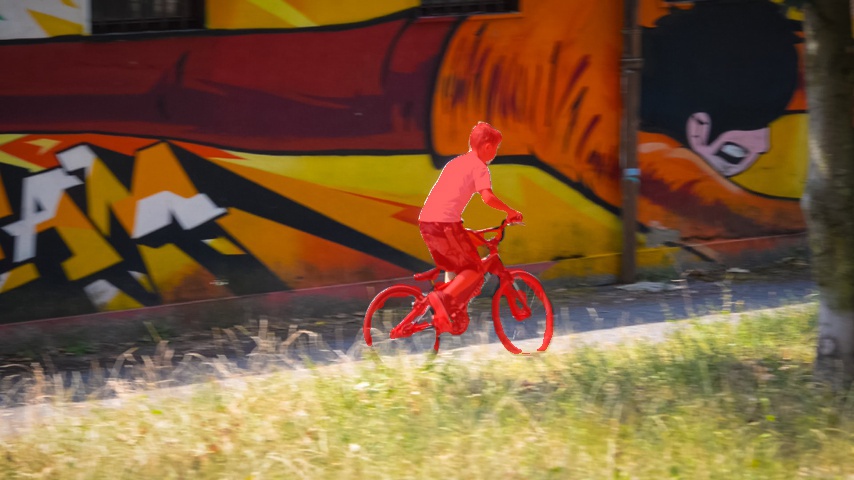}\hspace{1px}%
  \includegraphics[width=0.18\textwidth, frame]{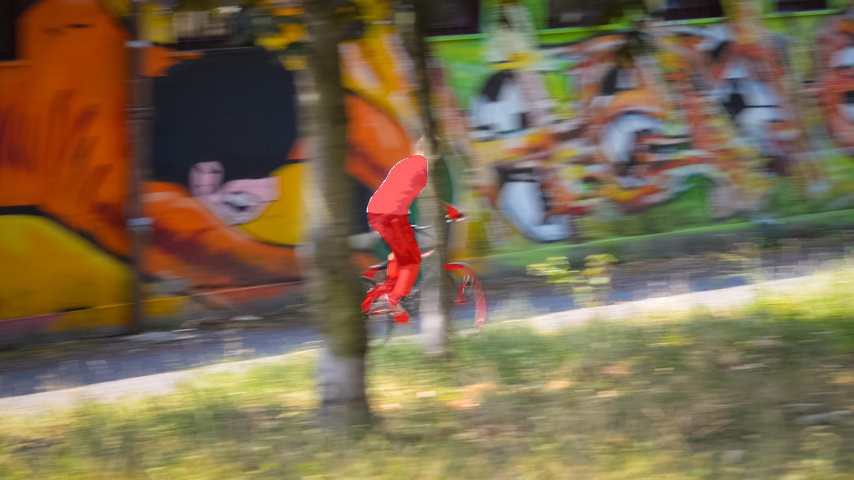}\hspace{1px}%
  \includegraphics[width=0.18\textwidth, frame]{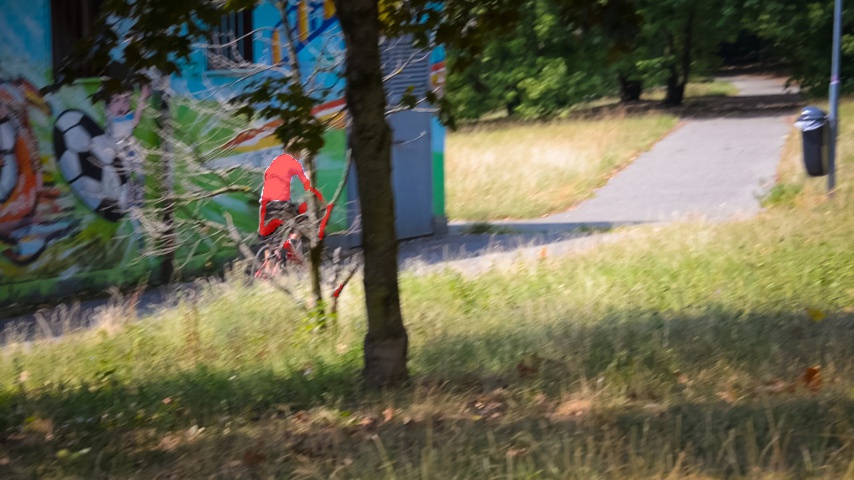}\hspace{1px}%
  \includegraphics[width=0.18\textwidth, frame]{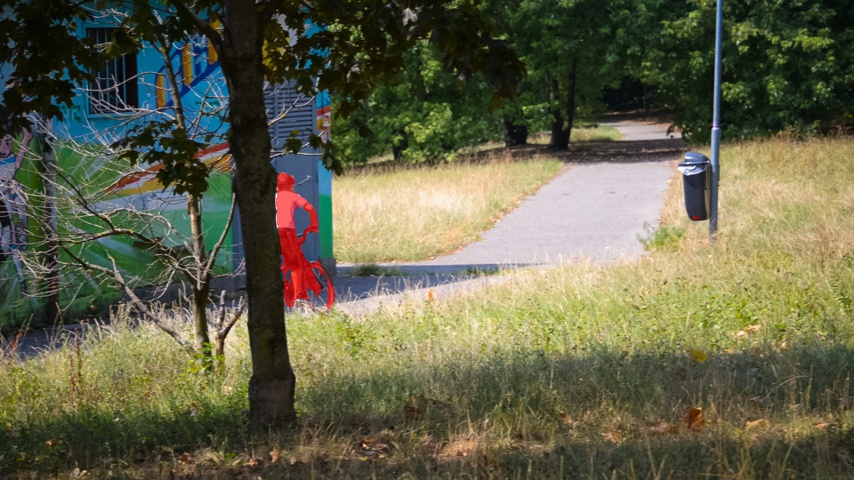}\hspace{1px}%
  \includegraphics[width=0.18\textwidth, frame]{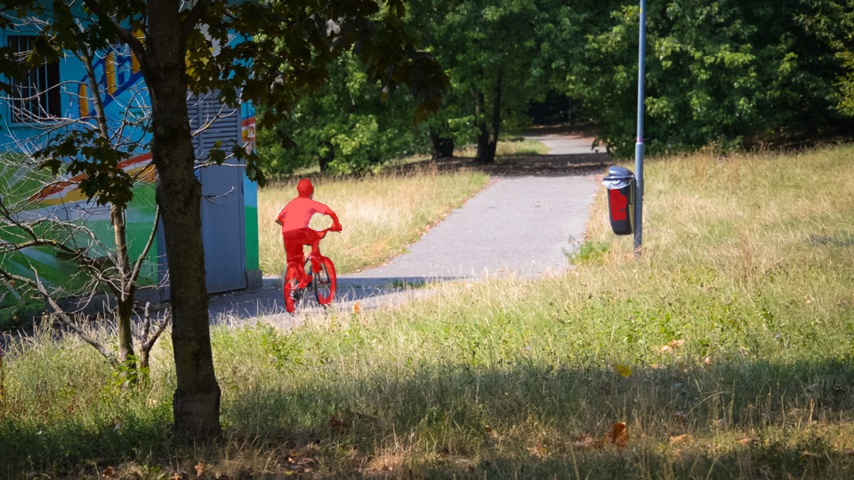}\\
  \includegraphics[width=0.18\textwidth, frame]{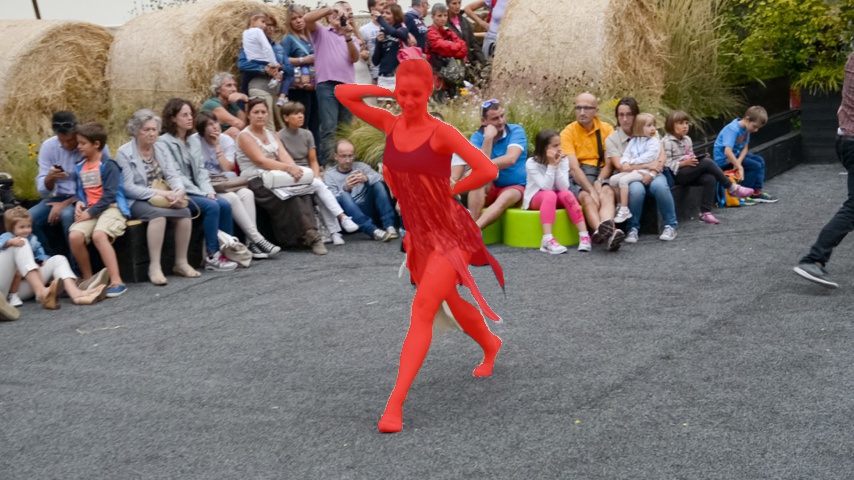}\hspace{1px}%
  \includegraphics[width=0.18\textwidth, frame]{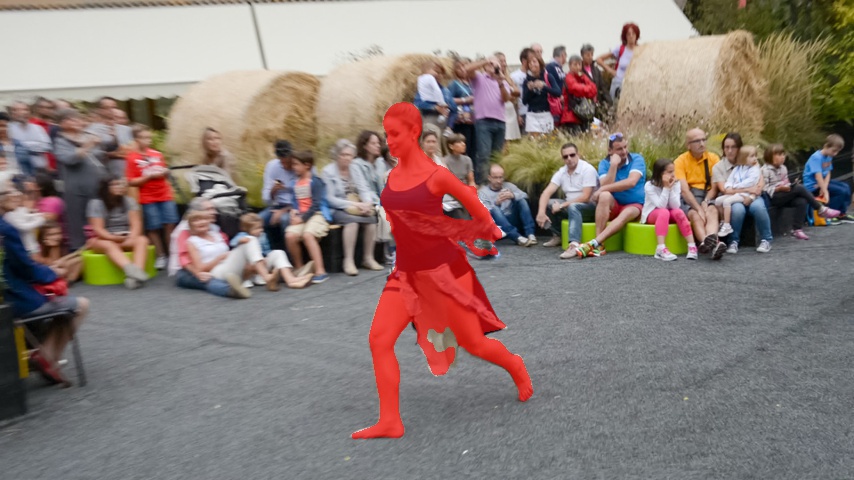}\hspace{1px}%
  \includegraphics[width=0.18\textwidth, frame]{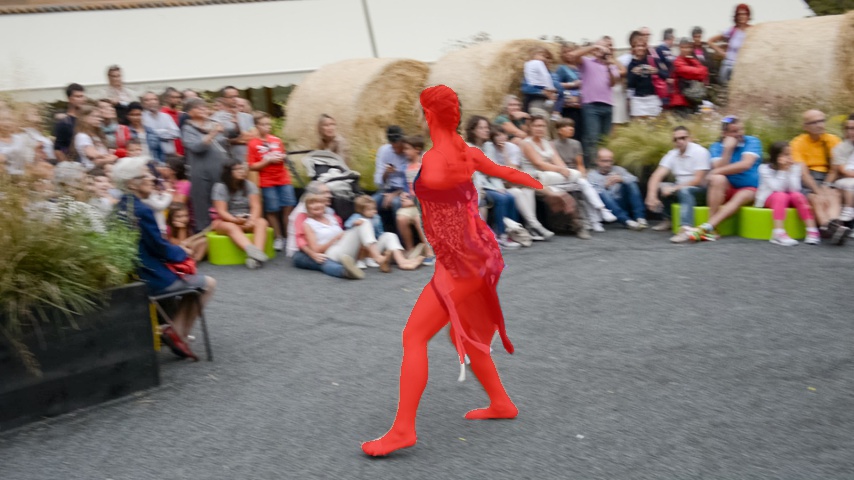}\hspace{1px}%
  \includegraphics[width=0.18\textwidth, frame]{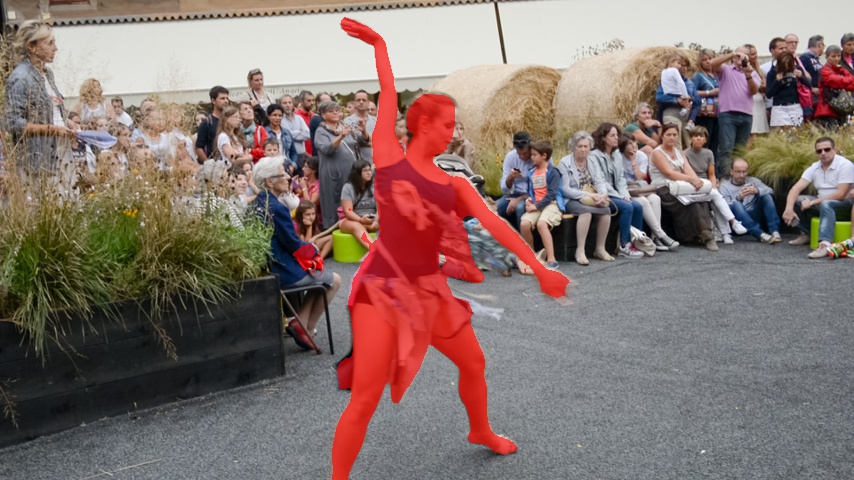}\hspace{1px}%
  \includegraphics[width=0.18\textwidth, frame]{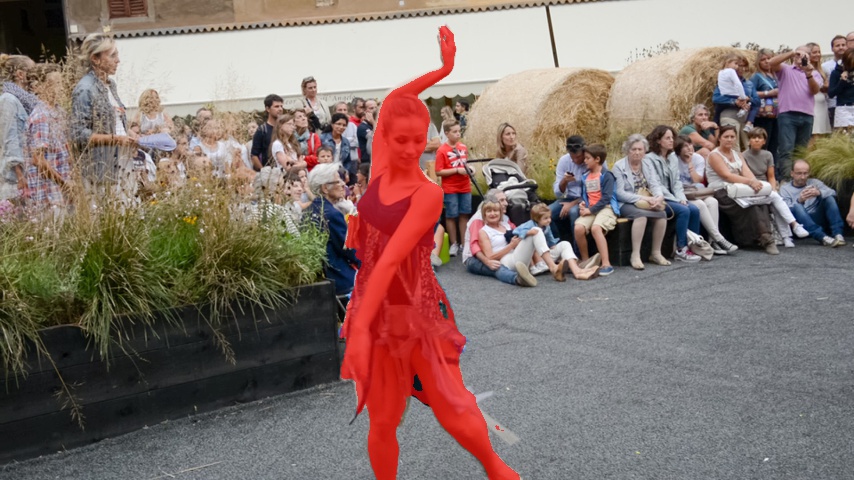}\\
  \caption{Additional Qualitative Results on DAVIS '16.}
 \end{figure}
 \vspace{-12pt}
 \begin{figure}[h]
\centering
  \includegraphics[width=0.18\textwidth, frame]{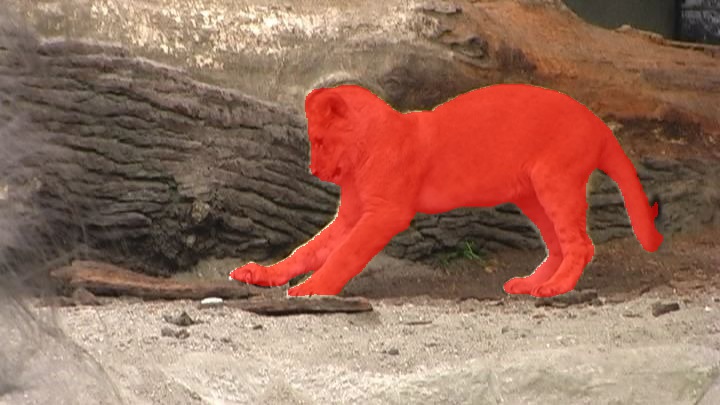}\hspace{1px}%
  \includegraphics[width=0.18\textwidth, frame]{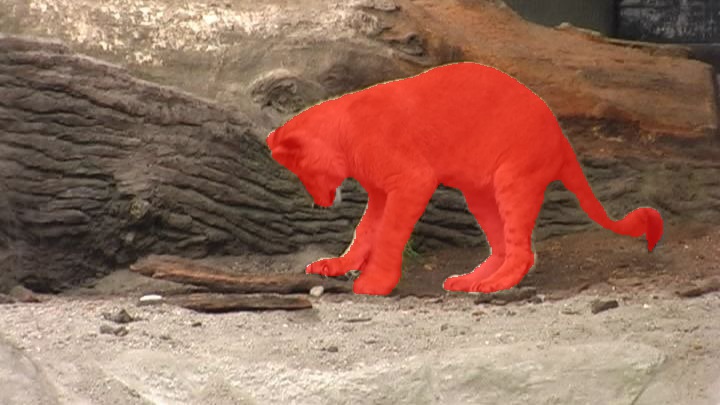}\hspace{1px}%
  \includegraphics[width=0.18\textwidth, frame]{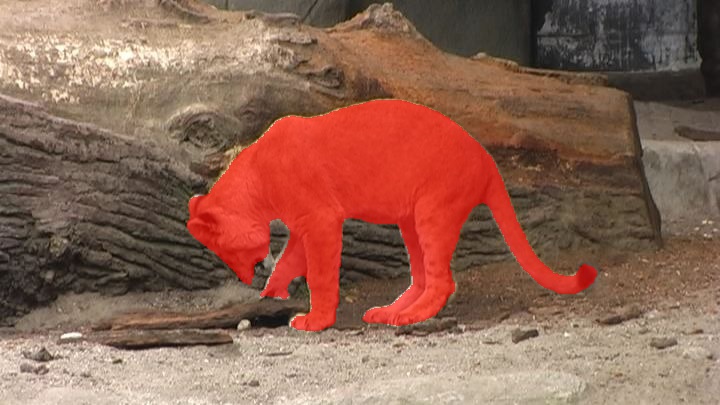}\hspace{1px}%
  \includegraphics[width=0.18\textwidth, frame]{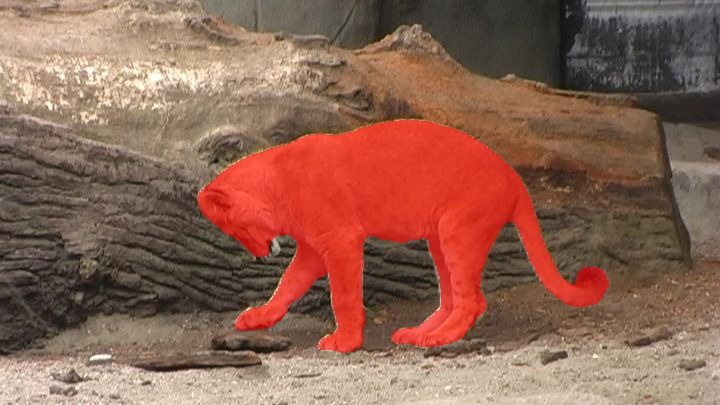}\hspace{1px}%
  \includegraphics[width=0.18\textwidth, frame]{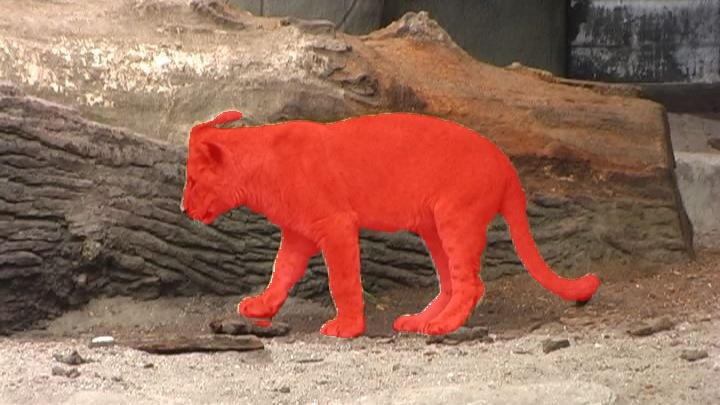}\\
  \includegraphics[width=0.18\textwidth, frame]{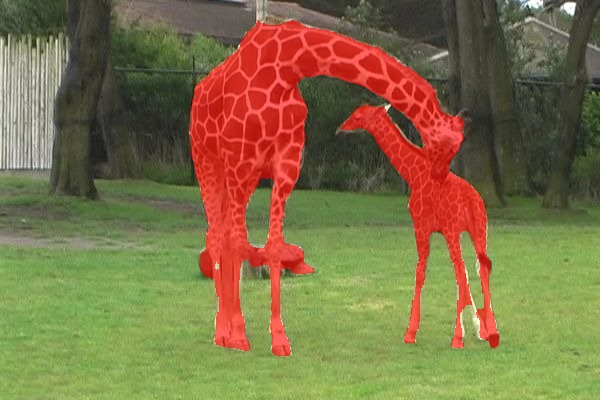}\hspace{1px}%
  \includegraphics[width=0.18\textwidth,frame]{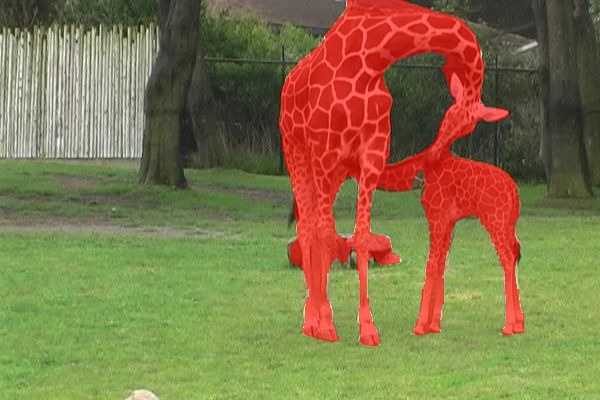}\hspace{1px}
  \includegraphics[width=0.18\textwidth, trim={0 0 0 0}, clip,frame]{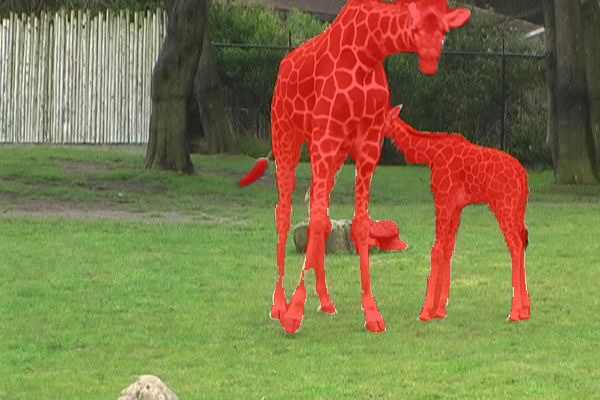}\hspace{1px}%
  \includegraphics[width=0.18\textwidth, trim={0 0 0 0}, clip,frame]{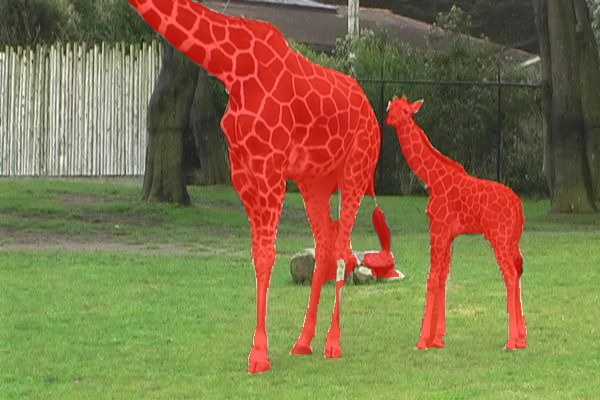}\hspace{1px}%
  \includegraphics[width=0.18\textwidth, trim={0 0 0 0}, clip,frame]{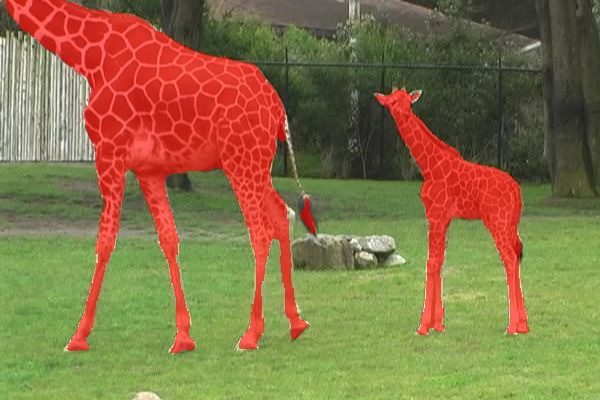}\\
  \includegraphics[width=0.18\textwidth, trim={0 30 0 0}, clip, frame]{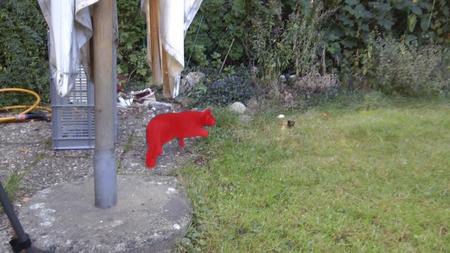}\hspace{1px}%
  \includegraphics[width=0.18\textwidth, trim={0 30 0 0}, clip, frame]{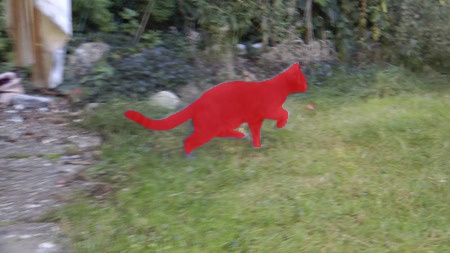}\hspace{1px}%
  \includegraphics[width=0.18\textwidth, trim={0 30 0 0}, clip, frame]{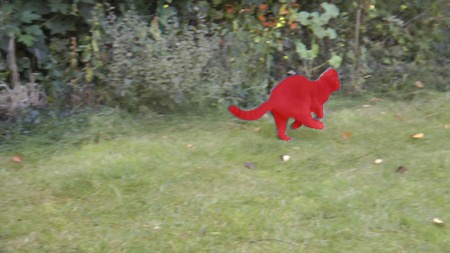}\hspace{1px}%
  \includegraphics[width=0.18\textwidth, trim={0 30 0 0}, clip, frame]{figures/supplementary/FMBS/cat06/080.jpg}\hspace{1px}%
  \includegraphics[width=0.18\textwidth, trim={0 30 0 0}, clip, frame]{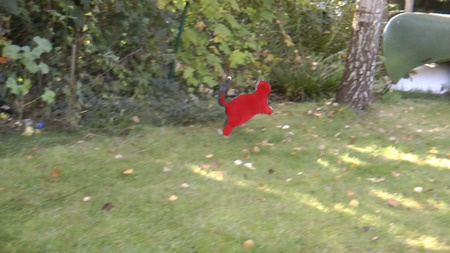}\\
  \caption{Qualitativate Results on FBMS Dataset.}
  \end{figure}
 \vspace{-15pt}
 \begin{figure}[h]
\centering
  \includegraphics[width=0.18\textwidth, trim={0 30 0 0}, clip, frame]{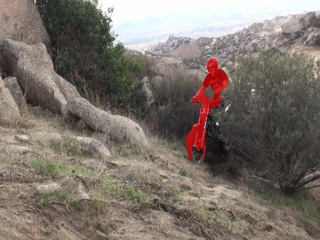}\hspace{1px}%
  \includegraphics[width=0.18\textwidth, trim={0 30 0 0}, clip, frame]{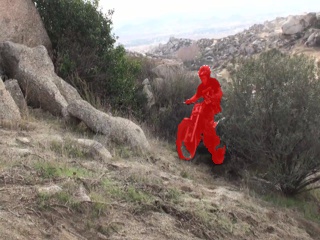}\hspace{1px}%
  \includegraphics[width=0.18\textwidth, trim={0 30 0 0}, clip, frame]{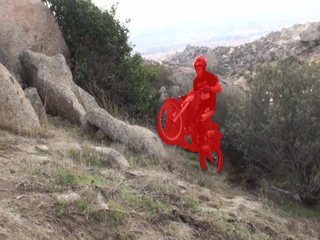}\hspace{1px}%
  \includegraphics[width=0.18\textwidth, trim={0 30 0 0}, clip, frame]{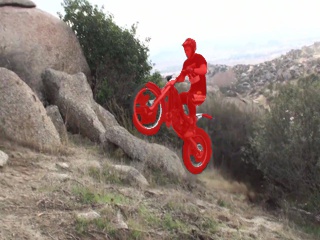}\hspace{1px}%
  \includegraphics[width=0.18\textwidth, trim={0 30 0 0}, clip, frame]{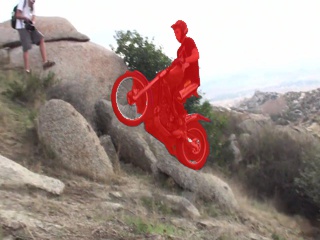}\\
  \includegraphics[width=0.18\textwidth, trim={0 0 0 50}, clip, frame]{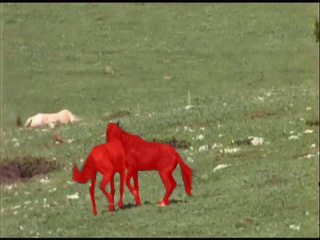}\hspace{1px}%
  \includegraphics[width=0.18\textwidth, trim={0 0 0 50}, clip, frame]{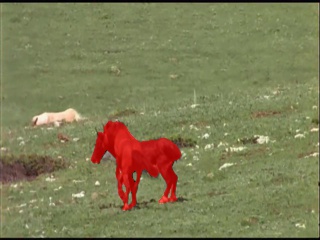}\hspace{1px}%
  \includegraphics[width=0.18\textwidth, trim={0 0 0 50}, clip, frame]{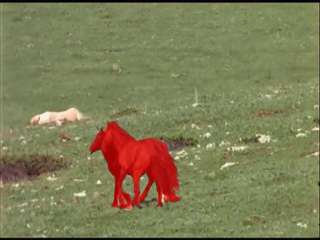}\hspace{1px}%
  \includegraphics[width=0.18\textwidth, trim={0 0 0 50}, clip, frame]{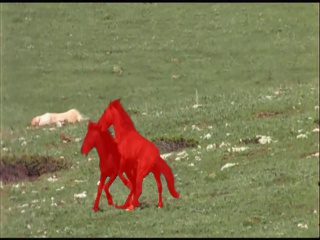}\hspace{1px}%
  \includegraphics[width=0.18\textwidth, trim={0 0 0 50}, clip, frame]{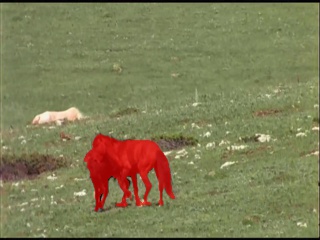}\\
  \includegraphics[width=0.18\textwidth, trim={0 30 0 0}, clip, frame]{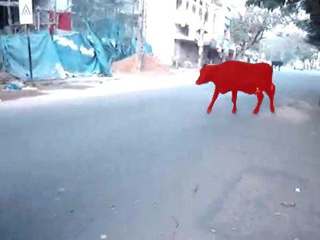}\hspace{1px}%
  \includegraphics[width=0.18\textwidth, trim={0 30 0 0}, clip, frame]{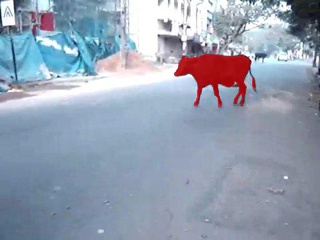}\hspace{1px}%
  \includegraphics[width=0.18\textwidth, trim={0 30 0 0}, clip, frame]{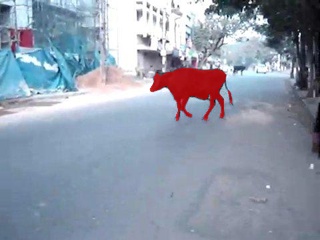}\hspace{1px}%
  \includegraphics[width=0.18\textwidth, trim={0 30 0 0}, clip, frame]{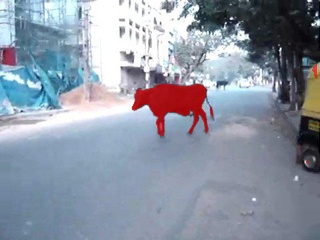}\hspace{1px}%
  \includegraphics[width=0.18\textwidth, trim={0 30 0 0}, clip, frame]{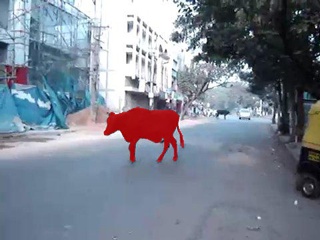}\\
  
\vspace{-2px}
  \caption{Qualitative Results on ViSal Dataset.}
  \label{fig:teaser}
\end{figure}

\end{document}